\documentclass[10pt,twocolumn,letterpaper]{article}
\pdfoutput=1
\usepackage{cvpr}
\usepackage{times}
\usepackage{epsfig}
\usepackage{graphicx}
\usepackage{amsmath}
\usepackage{amssymb}
\usepackage{xspace}

\makeatletter
\DeclareRobustCommand\onedot{\futurelet\@let@token\@onedot}
\def\@onedot{\ifx\@let@token.\else.\null\fi\xspace}
\def\eg{\emph{e.g}\onedot}

\def\etal{\emph{et al}\onedot}
\makeatother

\usepackage[breaklinks=true,bookmarks=false]{hyperref}

\cvprfinalcopy

\newcommand{\PAR}[1]{\vskip4pt \noindent {\bf #1~}}
\newcommand{\PARbegin}[1]{\noindent {\bf #1~}}

\begin{document}
\graphicspath{{images/}}

\title{How Robust is 3D Human Pose Estimation to Occlusion?}

\author{Istv\'{a}n S\'{a}r\'{a}ndi$^{1}$, Timm Linder$^{2}$, Kai O. Arras$^{2}$ and Bastian Leibe$^{1}$\vspace{8pt}\\
$^{1}$Visual Computing Institute, RWTH Aachen University \\ {\tt\small \{sarandi,leibe\}@vision.rwth-aachen.de} \\
$^{2}$Robert Bosch GmbH, Corporate Research \\ {\tt\small \{timm.linder,kaioliver.arras\}@de.bosch.com}%
}

\maketitle
\begin{abstract}
Occlusion is commonplace in realistic human-robot shared environments, yet its effects are not considered in standard 3D human pose estimation benchmarks. This leaves the question open: how robust are state-of-the-art 3D pose estimation methods against partial occlusions? We study several types of synthetic occlusions over the Human3.6M dataset and find a method with state-of-the-art benchmark performance to be sensitive even to low amounts of occlusion. Addressing this issue is key to progress in applications such as collaborative and service robotics. We take a first step in this direction by improving occlusion-robustness through training data augmentation with synthetic occlusions. This also turns out to be an effective regularizer that is beneficial even for non-occluded test cases.%
\end{abstract}

\section{Introduction}
\begin{figure}[t]
\centering
\includegraphics[scale=0.17]{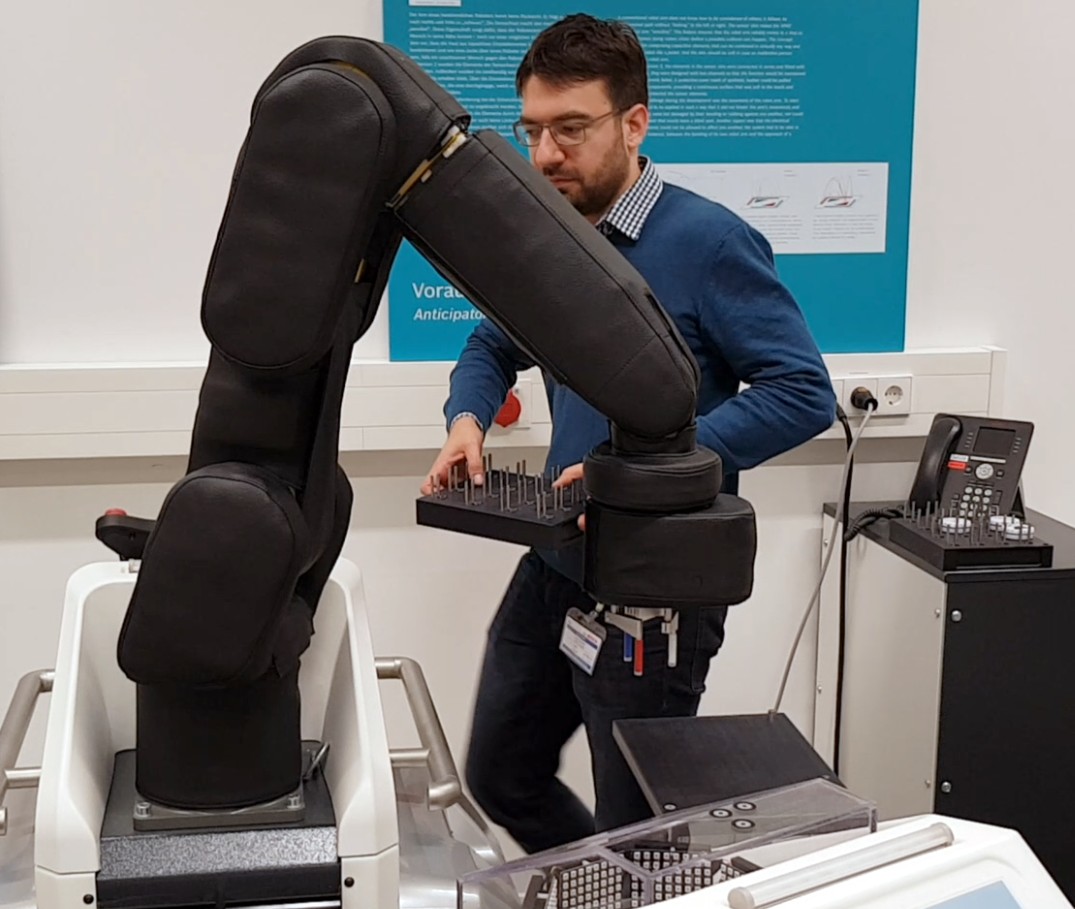}
\caption{Example of partial occlusions in the context of shared human-robot workspaces. Note how easily we humans can guess the rough pose of the person behind the occlusion. Can current 3D human pose estimation methods do that as well?}
\label{fig:robot_occlusion}
\end{figure}

To collaborate with humans and to understand their actions, collaborative and service robots need the ability to reason about human pose in 3D space. An important challenge in realistic environments is that humans are often only seen partially, \eg, standing behind machine parts or carrying objects in front of the body (see Fig.~\ref{fig:robot_occlusion}). Robust robotics solutions need to handle such disturbances gracefully and make use of the visual cues still present in the scene to reason around the occlusion.

Although recent years have brought significant advances in 3D human pose estimation, as measured on standard computer vision benchmarks such as Human3.6M~\cite{Ionescu14PAMI}\cite{Ionescu11ICCV}, the behavior of models under occlusion remains largely unexplored, as the benchmarks do not systematically model occlusion effects.

To our knowledge, we present the first systematic study of various types of test-time (synthetic) occlusions in 3D human pose estimation from a single RGB image. As we will see, ignoring the aspect of occlusions may cause model accuracy to rapidly deteriorate, even under mild occlusion levels, despite the good benchmark performance. Such sudden and unexpected failures in the robot's perception would prevent smooth and comfortable human-robot interaction and may lead to safety hazards. Furthermore, we demonstrate that simple occlusion data augmentation during training increases model robustness. This augmentation also improves performance even for non-occluded test images. Our approach is efficient and suitable for high frame-rate applications.

\section{Related Work}

\PARbegin{3D Human Pose Estimation.}
3D human pose estimation has seen rapid progress in recent years. For a thorough overview of approaches, we refer the reader to Sarafianos \etal's survey~\cite{Sarafianos16CVIU}. Current state-of-the-art methods use deep neural networks, either directly on the input image or on the output of a 2D pose estimator. Based on the sweeping success of heatmap-based representations in 2D human pose estimation (\eg, \cite{Newell16ECCV}), heatmaps have recently been also adopted in 3D methods, including volumetric~\cite{Pavlakos17CVPR}\cite{Sun17arXiv} and marginal heatmaps~\cite{Nibali18arXiv2}.

\PAR{Occlusions, Erasing and Copy-Pasting.}
In a pre-deep learning study based on silhouettes and HOG features, Huang \etal tackled occlusions in 3D pose estimation from RGB~\cite{Huang09ACCV}, but their analysis was limited to walking actions and occlusions with two rectangles. Occlusion effects have also been studied in 3D pose estimation from depth input~\cite{Rafi15CVPRW}, where exploiting semantic information from the occluder itself was found to improve predictions.

Data augmentation by erasing a rectangular block from the input has recently been concurrently investigated under the names \emph{Random Erasing}~\cite{Zhong17arXiv} and \emph{Cutout}~\cite{DeVries17arXiv}, for image classification, object detection, and person re-identification.
Similarly, synthetically placing objects into a scene by image-level \emph{copy-pasting} has been shown to help object detection~\cite{Dwibedi17ICCV}\cite{Dvornik18arXiv}\cite{Georgakis17arXiv}. However, those methods are trained to detect these pasted objects, while in our case the task is to infer what lies behind them. Ke \etal~\cite{Ke18arXiv} augment training images for 2D human pose estimation by copying background patches over some of the body joints. Research on facial landmark localization has investigated and modeled occlusions for a long time~\cite{Burgos13ICCV}\cite{Ghiasi14CVPR}, including augmenting training images with randomly pasted occluding objects~\cite{Yuen17TIV}.

\section{Approach}

\begin{figure}[tpb]
\centering
\includegraphics[scale=0.3]{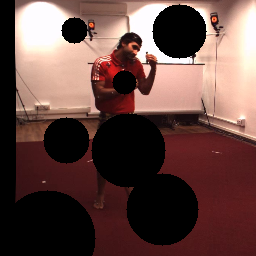} \hspace{1mm}
\includegraphics[scale=0.3]{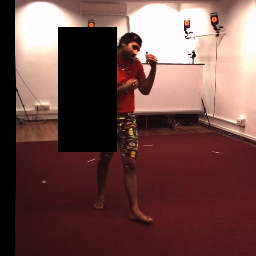} \vspace{2mm} \\
\includegraphics[scale=0.3]{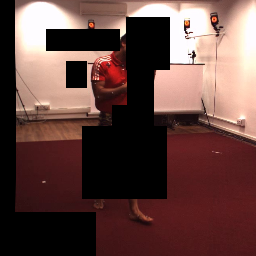} \hspace{1mm}
\includegraphics[scale=0.3]{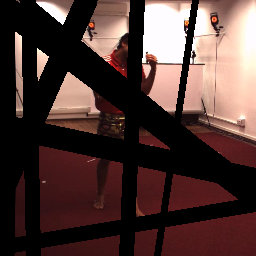}
\caption{Examples of the applied geometric occlusions: circles, a single rectangle~\cite{Zhong17arXiv}, rectangles, oriented bars. See Fig. \ref{fig:pred_change_examples} for an example with Pascal VOC objects.}
\label{fig:occlusion_types}
\end{figure}

In this paper we study the effect of occlusion on the accuracy of 3D human pose estimation. To this end, we have devised a 3D pose estimation approach that reaches state-of-the-art benchmark performance, leading us to expect that the observations drawn from our experiments also transfer to other models.

\PAR{Architecture.}
We use a fully convolutional net to predict volumetric body joint heatmaps from the input RGB image, based on a ResNet-50~\cite{He16CVPR} backbone architecture. 
After discarding the global average pooling layer, we adjust the number of output channels of the ResNet to be the product of the number of joints and the number of heatmap-voxels along the depth axis. Reshaping the resulting tensor yields the volumetric heatmaps. Nominal stride and depth discretization are configured to yield heatmaps of size $16\times 16 \times 16$ for an image of size $256 \times 256$.
Given the volumetric heatmap, coordinate predictions are obtained using soft-argmax~\cite{Levine16JMLR}\cite{Nibali18arXiv}\cite{Sun17arXiv}. As in \cite{Pavlakos17CVPR}, the $x$ and $y$ coordinates are interpreted as image space coordinates, while $z$ is the depth of the particular joint relative to the root (pelvis) joint depth, with the 16 voxels covering 2 meters. In order to concentrate on the aspect of articulated pose, as opposed to person localization, we assume that the true root joint depth is given by an oracle at test time. The coordinates are back-projected to camera space using the known camera intrinsics. Finally, the $L^1$ loss is computed on the predicted and ground truth 3D coordinates in camera space. Since all of the preceding operations are differentiable, the network can be trained end-to-end.

\section{Experimental Setup}

\begin{figure}[tpb]
\centering
\includegraphics[scale=0.1]{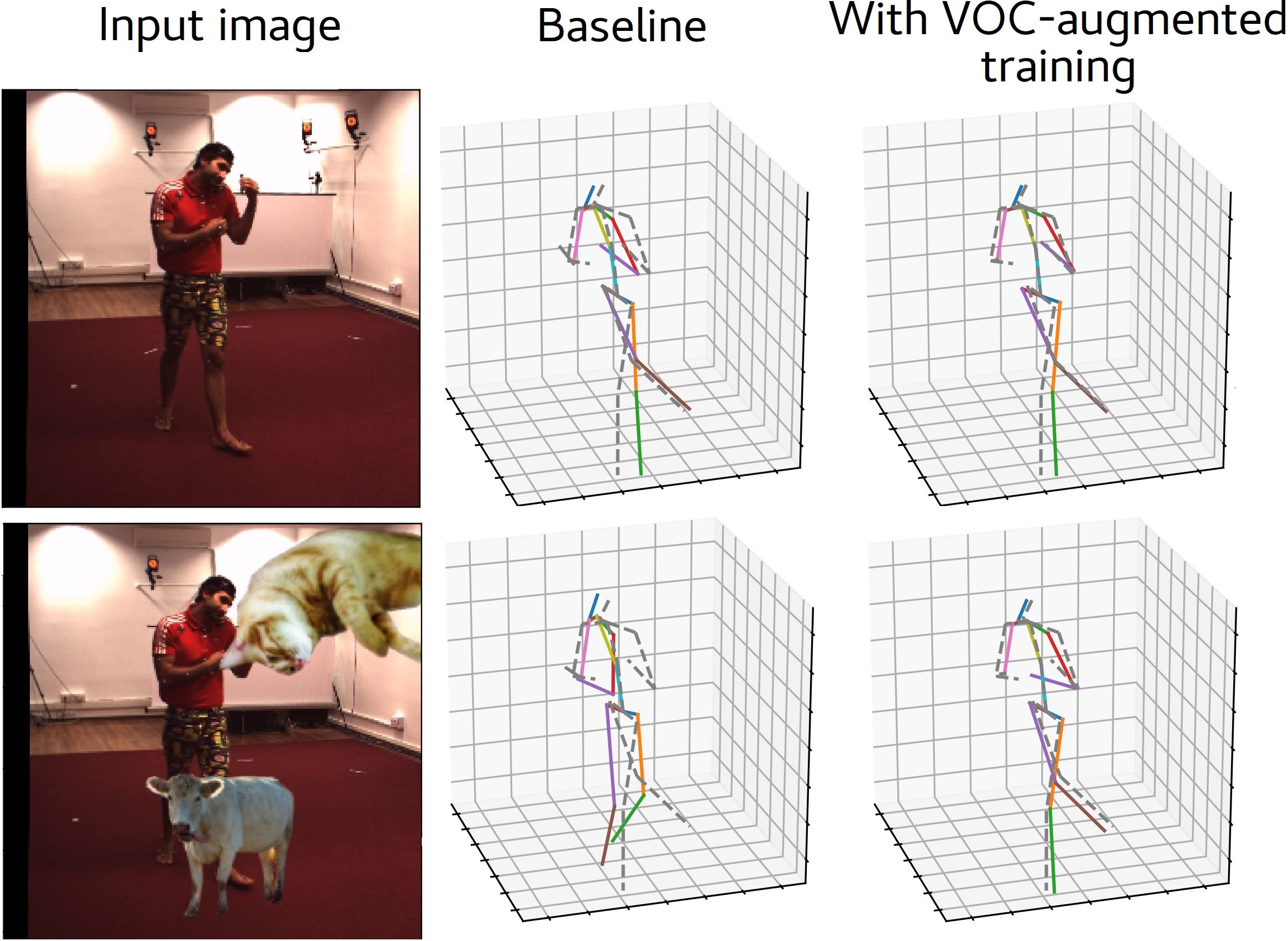}
\caption{Prediction change in the presence of synthetic test-time occlusion. Ground truth is shown with grey dashed lines, predictions with colorful ones. The baseline model fails to predict the pose of the occluded limbs, while the model trained with occlusion augmentation behaves more robustly.}
\label{fig:pred_change_examples}
\end{figure}

\setlength\tabcolsep{0.8mm}
\begin{table*}[t]
\footnotesize
\centering
\begin{tabular}{lrrrrrrrrrrrrrrrr}
\hline
                                  & Direct & Discuss    & Eat & Greet & Phone & Photo & Pose  & Purch. & Sit   & SitD  & Smoke & Wait  & Walk & WalkD &WalkT & Avg \\
\hline
Zhou \cite{Zhou16CVPR}     & 87.4  & 109.3 & 87.1 & 103.2 & 116.2 & 139.5 & 106.9 & 99.8  & 124.5 & 199.2 & 107.4 & 118.1 & 79.4 & 114.2 & 97.7 & 113.0 \\
Tekin \cite{Tekin16CVPR}       & 102.4 & 147.7 & 88.8 & 125.4 & 118.0 & 182.7 & 112.4 & 129.2 & 138.9 & 224.9 & 118.4 & 138.8 & 55.1 & 126.3 & 65.8 & 125.0 \\
Zhou \cite{Zhou16ECCV}         & 91.8  & 102.4 & 97.0 & 98.8  & 113.4 & 125.2 & 90.0  & 93.8  & 132.2 & 159.0 & 106.9 & 94.4  & 79.0 & 126.0 & 99.0 & 107.3 \\
Sun \cite{Sun17ICCV}   & 90.2  & 95.5  & 82.3 & 85.0  &  87.1 & 94.5  & 87.9  & 93.4  & 100.3 & 135.4 & 91.4  & 87.3  & 78.0 & 90.4  & 86.5 & 92.4  \\
Sun \cite{Sun17arXiv} (ArXiv)                   & 63.8  & 64.0  & 56.9 & 64.8  &  62.1 & 70.4  & 59.8  & 60.1  &  71.6 &  91.7 & 60.9  &  65.1 & 51.3 & 63.2  & 55.4 & 64.1  \\
Pavlakos \cite{Pavlakos17CVPR} & 67.4  & 72.0  & 66.7 & 69.1  &  72.0 & 77.0  & 65.0  & 68.3  &  83.7 &  96.5 & 71.7  &  65.8 & 59.1 & 74.9  & 63.2 & 71.9 \\
\hline
Pavlakos \cite{Pavlakos17CVPR} (known root depth) &  59.3 & 64.9 & 59.4 & 61.3 & 65.1 & 69.0 & 57.1 & 60.1 & 75.1 & 91.9 & 64.5 & 59.6 & 66.8 & 53.7 & 56.8 & 64.8  \\
Ours (no occlusion augm.)                    &  60.2 &  64.1 &  55.9 &  58.3 &  63.8 &  69.5 &  58.8 &  64.4 &  67.7 &  90.8 &  61.9 &  59.2 &  66.0 &  56.9 &  50.8 & 63.3  \\
\hspace{2mm} w/ circles augm. & 52.9 &  58.0 &  51.8 &  54.8 &  56.9 &  62.6 &  51.4 &  55.0 &  64.7 &  79.2 &  56.3 &  52.5 &  58.8 &  47.9 &  43.0 & 56.8  \\
\hspace{2mm} w/ single rectangle augm.\cite{Zhong17arXiv} & 52.0 &  58.6 &  51.0 &  53.5 &  56.1 &  62.6 &  51.5 &  54.2 &  65.7 &  71.2 &  56.1 &  52.9 &  58.2 &  47.8 &  42.9 & 56.1 \\
\hspace{2mm} w/ rectangles augm. & 51.9 &  57.9 &  52.5 &  54.2 &  57.3 &  61.9 &  51.7 &  55.2 &  63.4 &  76.7 &  56.5 &  51.7 &  58.8 &  47.8 &  43.4 & 56.5  \\
\hspace{2mm} w/ bars augm. & 55.0 &  60.1 &  54.1 &  56.4 &  59.9 &  64.9 &  52.4 &  59.5 &  67.7 &  88.7 &  58.5 &  54.2 &  62.4 &  50.0 &  45.4 & 59.6 \\
\hspace{2mm} w/ VOC objects augm. & 51.2 &  58.7 &  51.7 &  53.4 &  56.8 &  59.3 &  50.7 &  52.6 &  65.5 &  73.2 &  56.8 &  51.4 &  56.6 &  47.0 &  42.4 & 55.8 \\
\hspace{2mm} w/ mixture augm. & 51.3 &  57.8 &  52.5 &  53.8 &  55.9 &  58.7 &  50.9 &  52.8 &  66.7 &  77.1 &  56.6 &  51.7 &  56.6 &  47.6 &  42.8 & 56.1  \\
\hline \\
\end{tabular}
\caption{Mean per joint position error on Human3.6M for methods using no extra pose datasets in training. Methods below the line have access to the ground-truth root joint depth at test-time. (No synthetic occlusions are used on the test inputs.)}
\label{tab:comparison_prior_work}
\end{table*}

\PAR{Dataset.}
Human3.6M~\cite{Ionescu14PAMI} is the largest public 3D pose estimation dataset. It contains 11 subjects imitating 15 actions in a controlled indoor environment while being recorded with 4 cameras and a motion capture system. Following the most common experimental protocol in the literature, we use five subjects (S1, S5, S6, S7, S8) for training and two (S9, S11) for testing. We train action-agnostic models, as opposed to action-specific ones.

\PAR{Dataset Subsampling.}
To reduce the redundancy in training poses, we adaptively subsample the frames similarly to \cite{Mehta17TOG}, only keeping a frame when at least one body joint moves by at least 30 mm compared to the last kept frame. For the test set we follow prior work and use every 64\textsuperscript{th} frame.

\PAR{Image Preprocessing.}
Before feeding an image to the network, we center and zoom it on the person, at a resolution of $256\times 256$ px. To ensure correct perspective (with the principal point at the image center), we reproject the image onto a virtual camera pointed at the center of the person's bounding box, as provided in the dataset. Scaling is applied so that the larger side of the person's bounding box covers about 80\% of the image side length. Common data augmentation techniques are used in training, including random rotation, scaling, translation, horizontal flipping, as well as image filtering such as color distortions and blurs.

\PAR{Evaluation Metrics.}
Following standard practice on Human3.6M, we evaluate prediction accuracy by the so-called mean per joint position error (MPJPE), which is the mean Euclidean error of all joints after skeleton alignment at the root (pelvis) joint. Procrustes alignment is not used.

\PAR{Synthetic Occlusions for Robustness Analysis.}
We consider solid black shapes and some more realistic object segments from the Pascal VOC 2012 dataset~\cite{Everingham11} as occluders in this study (see Fig. \ref{fig:occlusion_types} and \ref{fig:pred_change_examples}). The number, position and size of the objects are generated at random.
We define the \emph{degree of occlusion} as the percentage of occluded pixels inside the person's bounding box and vary this quantity between 0\% and 70\%.

\PAR{Occlusion-Augmented Training.} We hypothesize that synthetic occlusion data augmentation during training can improve test-time occlusion-robustness. To verify this, we use the same kinds of occlusions as described in the previous section, with an additional \emph{mixture} variant, which uses one of the other types at random for each frame. We make sure to strictly separate the VOC objects used for training and testing. Furthermore, we try the RE-0 variant of Zhong \etal's \emph{random erasing}~\cite{Zhong17arXiv}, generating a single occluding black rectangle of random size according to their pseudo-code. We refer to this mode as \emph{single rectangle} in this paper.

To make these strategies comparable, we parameterize them such that the distribution of the number of occluded pixels is similar. Notably, we only apply these augmentations with 50\% probability for each frame. This was found important in prior work on occlusion augmentation~\cite{DeVries17arXiv}.

\PAR{Implementation Details.}
We use the implementation of ResNet-50v1 and the corresponding ImageNet-pretrained initial weights from the TensorFlow-Slim library~\cite{Silberman17Github}. Training is done with the Adam optimizer and a mini-batch size of 64, for 40 epochs, taking approximately 24 hours on an NVIDIA GeForce Titan X (Pascal) GPU.

\begin{figure}[tpb]
\centering
\includegraphics[width=245pt]{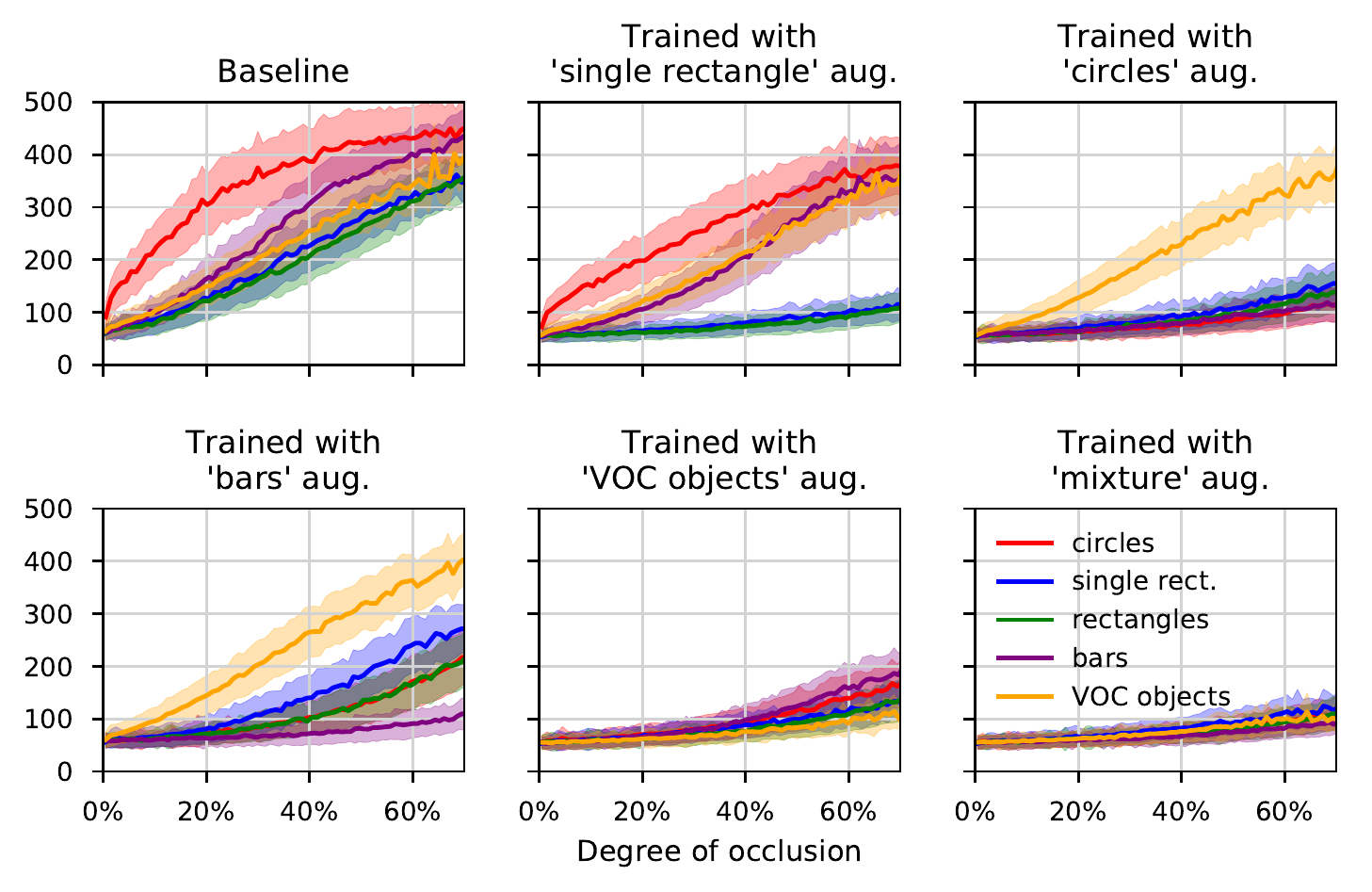}
\caption{Assessing occlusion-robustness on Human3.6M. Each subplot shows the performance when training with a particular augmentation method. Within a subplot, each line shows the mean and standard deviation of MPJPE under increasing degrees of occlusion of a particular type.}
\label{fig:occlusion_plots}
\end{figure}

\section{Results}

We start presenting our results by showing that our baseline model has state-of-the-art performance. We then show how performance deteriorates with test-time occlusions and that this can be mitigated using occlusion data augmentation. The augmentations are then shown to help even when the test images do not contain synthetic occlusions.

\PAR{Baseline Performance.}
The current state-of-the-art among published methods which use no extra 2D pose datasets for training is by Pavlakos \etal~\cite{Pavlakos17CVPR} (see Table \ref{tab:comparison_prior_work}). Since our evaluation assumes knowledge of the root joint depth at test time, we compare with Pavlakos \etal's performance under the same conditions, for which the results can be found in their supplementary material. Our baseline's MPJPE of 63.3 mm is already better than Pavlakos \etal's 64.8.

Sun \etal's (unpublished) method achieves an MPJPE of 64.1 mm~\cite{Sun17arXiv}, but it is unclear whether they use the known root joint depth or resolve scale ambiguity by other means.

\PAR{Robustness Analysis under Occlusion.}
We evaluate the robustness of our baseline model using different degrees and types of occlusions (see the top left plot of Fig.~\ref{fig:occlusion_plots}). We observe that circular occlusions cause by far the largest increase in error, the reason for which needs further investigation. Occlusions with oriented bars, VOC objects and rectangles lead to comparable performance loss. We note that rectangles are the least problematic type of occlusion, despite being a widely used test case in the literature.

Fig. \ref{fig:pred_change_examples} visualizes an example. The baseline network gives good predictions for the unoccluded case, but when we paste two Pascal VOC objects onto the image, prediction visibly fails for the affected limbs.

\begin{figure}[tpb]
\centering
\includegraphics[width=245pt]{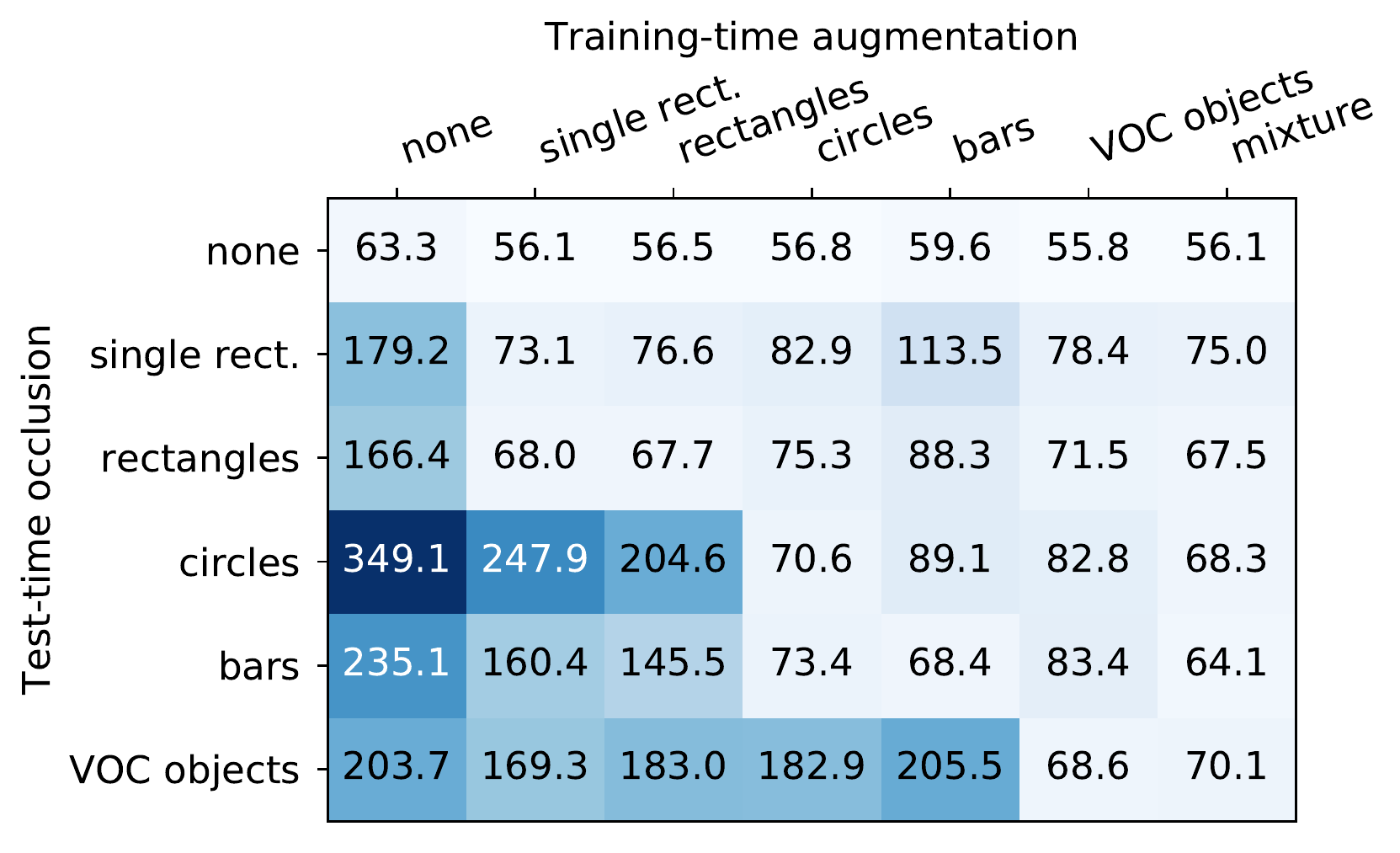}
\caption{Exploring how much each type of training-time data augmentation protects against each type of test occlusions. The numbers are the MPJPE averaged for degrees of occlusion between 10\% and 50\%.}
\label{fig:occ_train_test_pairs}
\end{figure}

\PAR{Augmentation Improves Occlusion-Robustness.}
We now turn to the evaluation of occlusion augmentation at training time for increased test-time occlusion-robustness. Fig. \ref{fig:occlusion_plots} and \ref{fig:occ_train_test_pairs} show the results. Erasing a single rectangle (as in \cite{Zhong17arXiv}) results in robustness against multiple rectangles at test time, but is much less effective for the other types of occlusions, being most sensitive to circles. Using several rectangles during training works slightly better than single-rectangle random erasing, but it, too, has difficulty in generalizing to other types of occlusion structures. Circular occlusion augmentation generalizes to all other simple geometric occlusion shapes, but barely helps when more realistic VOC objects are used as occluders at test time. VOC-augmentation, however, does generalize to both simple geometric shapes and other VOC objects (the objects used in training and testing are strictly separated). The qualitative difference in robustness when using this augmentation type is illustrated in Fig.~\ref{fig:pred_change_examples}. The network learned to use context cues and gives good prediction even for the almost fully-occluded lower left leg.
Finally, the combination of all these strategies proves to be effective against all of the analyzed occlusion types together.

\PAR{The Regularizing Effect of Occlusion Augmentation.}
In the previous section we have seen that training-time occlusion augmentation is helpful when evaluating on occluded test examples. Let us now look at the effect of these augmentation schemes when evaluating on the original test data without synthetic occlusions (see Table \ref{tab:comparison_prior_work}). All occlusion augmentation strategies are found to improve upon the baseline result, with the \emph{VOC objects} performing the best and \emph{bars} the worst.

\PAR{Runtime.}
Inference of the whole pipeline runs at 64, 165, and 204 fps for batch sizes of 1, 8, and 64 images, respectively, on a single NVIDIA GeForce Titan X (Pascal) GPU. This makes the method suitable for high frame rate applications.

\section{Conclusion}

We presented a systematic study of occlusion effects on 3D human pose estimation from a single RGB image, using an efficient ResNet-based test model. We found that despite producing state-of-the-art benchmark results, the network's performance quickly drops when synthetic occlusions are added. Circular structures turned out to be particularly problematic, the reason of which needs further study. We then showed that training-time occlusion data augmentation is effective in reducing occlusion-induced errors, while also improving the performance without test-time occlusions.

Future experiments should also target other datasets besides Human3.6M and it remains to be seen how well our findings about synthetic occlusions generalize to real ones.

\section*{Acknowledgments}

This project has been funded by a grant from the Bosch Research Foundation and by ILIAD (H2020-ICT-2016-732737).


{\small
\bibliographystyle{ieee}
\bibliography{abbrev_short,references}

\begin{thebibliography}{10}\itemsep=-1pt

\bibitem{Burgos13ICCV}
X.~P. Burgos-Artizzu, P.~Perona, and P.~Doll{\'a}r.
\newblock Robust face landmark estimation under occlusion.
\newblock In {\em ICCV}, 2013.

\bibitem{Ionescu11ICCV}
C.~S. Catalin~Ionescu, Fuxin~Li.
\newblock Latent structured models for human pose estimation.
\newblock In {\em ICCV}, 2011.

\bibitem{DeVries17arXiv}
T.~DeVries and G.~W. Taylor.
\newblock Improved regularization of convolutional neural networks with cutout.
\newblock {\em arXiv:1708.04552}, 2017.

\bibitem{Dvornik18arXiv}
N.~Dvornik, J.~Mairal, and C.~Schmid.
\newblock Modeling visual context is key to augmenting object detection
  datasets.
\newblock {\em arXiv:1807.07428}, 2018.

\bibitem{Dwibedi17ICCV}
D.~Dwibedi, I.~Misra, and M.~Hebert.
\newblock Cut, paste and learn: Surprisingly easy synthesis for instance
  detection.
\newblock In {\em ICCV}, 2017.

\bibitem{Everingham11}
M.~Everingham and J.~Winn.
\newblock The pascal visual object classes challenge 2012 (voc2012) development
  kit.
\newblock {\em Pattern Analysis, Statistical Modelling and Computational
  Learning, Tech. Rep}, 2011.

\bibitem{Georgakis17arXiv}
G.~Georgakis, A.~Mousavian, A.~C. Berg, and J.~Kosecka.
\newblock Synthesizing training data for object detection in indoor scenes.
\newblock {\em arXiv:1702.07836}, 2017.

\bibitem{Ghiasi14CVPR}
G.~Ghiasi and C.~C. Fowlkes.
\newblock Occlusion coherence: Localizing occluded faces with a hierarchical
  deformable part model.
\newblock In {\em CVPR}, 2014.

\bibitem{He16CVPR}
K.~He, X.~Zhang, S.~Ren, and J.~Sun.
\newblock Deep residual learning for image recognition.
\newblock In {\em CVPR}, 2016.

\bibitem{Huang09ACCV}
J.-B. Huang and M.-H. Yang.
\newblock Estimating human pose from occluded images.
\newblock In {\em ACCV}, 2009.

\bibitem{Ionescu14PAMI}
C.~Ionescu, D.~Papava, V.~Olaru, and C.~Sminchisescu.
\newblock Human3.6m: Large scale datasets and predictive methods for 3d human
  sensing in natural environments.
\newblock {\em PAMI}, 2014.

\bibitem{Ke18arXiv}
L.~Ke, M.-C. Chang, H.~Qi, and S.~Lyu.
\newblock Multi-scale structure-aware network for human pose estimation.
\newblock {\em arXiv:1803.09894}, 2018.

\bibitem{Levine16JMLR}
S.~Levine, C.~Finn, T.~Darrell, and P.~Abbeel.
\newblock End-to-end training of deep visuomotor policies.
\newblock {\em JMLR}, 17(1):1334--1373, 2016.

\bibitem{Mehta17TOG}
D.~Mehta et~al.
\newblock Vnect: Real-time 3d human pose estimation with a single rgb camera.
\newblock {\em ACM Trans. Graphics}, 36(4):44, 2017.

\bibitem{Newell16ECCV}
A.~Newell, K.~Yang, and J.~Deng.
\newblock Stacked hourglass networks for human pose estimation.
\newblock In {\em ECCV}, 2016.

\bibitem{Nibali18arXiv2}
A.~Nibali, Z.~He, S.~Morgan, and L.~Prendergast.
\newblock 3d human pose estimation with 2d marginal heatmaps.
\newblock {\em arXiv:1806.01484}, 2018.

\bibitem{Nibali18arXiv}
A.~Nibali, Z.~He, S.~Morgan, and L.~Prendergast.
\newblock Numerical coordinate regression with convolutional neural networks.
\newblock {\em arXiv:1801.07372}, 2018.

\bibitem{Pavlakos17CVPR}
G.~Pavlakos, X.~Zhou, K.~G. Derpanis, and K.~Daniilidis.
\newblock Coarse-to-fine volumetric prediction for single-image 3d human pose.
\newblock In {\em CVPR}, 2017.

\bibitem{Rafi15CVPRW}
U.~Rafi, J.~Gall, and B.~Leibe.
\newblock A semantic occlusion model for human pose estimation from a single
  depth image.
\newblock In {\em CVPR Workshops}, 2015.

\bibitem{Sarafianos16CVIU}
N.~Sarafianos, B.~Boteanu, B.~Ionescu, and I.~A. Kakadiaris.
\newblock 3d human pose estimation: A review of the literature and analysis of
  covariates.
\newblock {\em CVIU}, 152:1--20, 2016.

\bibitem{Silberman17Github}
N.~Silberman and S.~Guadarrama.
\newblock Tensorflow-slim image classification model library.
\newblock https://github.com/tensorflow/models/tree/master/research/slim, 2016.

\bibitem{Sun17ICCV}
X.~Sun, J.~Shang, S.~Liang, and Y.~Wei.
\newblock Compositional human pose regression.
\newblock In {\em ICCV}, 2017.

\bibitem{Sun17arXiv}
X.~Sun, B.~Xiao, S.~Liang, and Y.~Wei.
\newblock Integral human pose regression.
\newblock {\em arXiv:1711.08229}, 2017.

\bibitem{Tekin16CVPR}
B.~Tekin, A.~Rozantsev, V.~Lepetit, and P.~Fua.
\newblock Direct prediction of 3d body poses from motion compensated sequences.
\newblock In {\em CVPR}, 2016.

\bibitem{Yuen17TIV}
K.~Yuen and M.~M. Trivedi.
\newblock An occluded stacked hourglass approach to facial landmark
  localization and occlusion estimation.
\newblock {\em IEEE Trans. Intel. Veh.}, 2(4):321--331, 2017.

\bibitem{Zhong17arXiv}
Z.~Zhong, L.~Zheng, G.~Kang, S.~Li, and Y.~Yang.
\newblock Random erasing data augmentation.
\newblock {\em arXiv:1708.04896}, 2017.

\bibitem{Zhou16ECCV}
X.~Zhou, X.~Sun, W.~Zhang, S.~Liang, and Y.~Wei.
\newblock Deep kinematic pose regression.
\newblock In {\em ECCV}, 2016.

\bibitem{Zhou16CVPR}
X.~Zhou, M.~Zhu, S.~Leonardos, K.~G. Derpanis, and K.~Daniilidis.
\newblock Sparseness meets deepness: 3d human pose estimation from monocular
  video.
\newblock In {\em CVPR}, 2016.

\end{thebibliography}
}

\end{document}